\documentclass[11pt]{article}

\usepackage[a4paper,margin=1in]{geometry}
\usepackage[T1]{fontenc}
\usepackage[utf8]{inputenc}
\usepackage{lmodern}
\usepackage{microtype}
\usepackage{graphicx}
\usepackage{amsmath,amssymb,amsfonts}
\usepackage{xcolor}
\usepackage{booktabs}
\usepackage{url}
\usepackage{caption}
\usepackage{subcaption}
\usepackage{enumitem}
\usepackage{placeins}
\usepackage{float}
\usepackage{authblk}
\usepackage[backend=biber,style=numeric,sorting=none,maxbibnames=99]{biblatex}
\usepackage[colorlinks=true,allcolors=blue]{hyperref}

\setlist{leftmargin=*,itemsep=2pt,topsep=3pt}
\hypersetup{
  pdfauthor={Athanasios Angelakis and Marta Gomez-Barrero},
  pdftitle={Compact Vision Models for Iris Presentation Attack Detection under Presentation Attack Instrument Shift and Environmental Degradation},
  pdfsubject={Iris presentation attack detection},
  pdfkeywords={iris presentation attack detection, PAD, presentation attack instrument, compact vision models, biometric recognition security}
}

\title{\textbf{Compact Vision Models for Iris Presentation Attack Detection under Presentation Attack Instrument Shift and Environmental Degradation}}

\author[1,2]{Athanasios Angelakis\thanks{ORCID: \href{https://orcid.org/0000-0003-1226-9560}{0000-0003-1226-9560}}}
\author[1]{Marta Gomez-Barrero\thanks{ORCID: \href{https://orcid.org/0000-0003-4581-5353}{0000-0003-4581-5353}}}
\affil[1]{BioML Lab, RI CODE, UniBw, Munich, Germany}
\affil[2]{EDS, Amsterdam UMC, University of Amsterdam, Amsterdam, Netherlands}
\affil[ ]{\texttt{athanasios.angelakis@unibw.de} \quad \texttt{marta.gomez-barrero@unibw.de}}
\date{}

\begin{document}

\maketitle
\begin{center}
\small Accepted at BIOSIG 2026. This preprint includes minor nomenclature and editorial corrections.
\end{center}
\vspace{0.5em}

\begin{abstract}
Iris presentation attack detection (PAD) is security-critical when a subsystem that appears reliable during development encounters presentation attack instruments (PAIs) or acquisition conditions absent from validation data. We benchmark three compact scratch-trained computer-vision models, each with at most approximately 0.26 million trainable parameters, on the Notre Dame subset of LivDet-Iris 2017 under PAI-driven domain shift and environmental degradation. All models are trained without external pretraining or data augmentation and evaluated over five seeds. A validation-selected threshold is transferred unchanged to the known-attack, unknown-attack, corrupted, and pooled test partitions. From known to unknown attack presentations, Attack Presentation Classification Error Rate (APCER) increases by 17.11-30.47 percentage points and Detection Equal Error Rate (D-EER) increases by 7.38-12.73 percentage points. At the validation-selected threshold, ZACH-ViT obtains the lowest unknown-attack APCER ($47.69 \pm 4.84\%$) and D-EER ($38.87 \pm 0.93\%$), while Compact-TransMIL obtains the lowest Bona Fide Presentation Classification Error Rate (BPCER). ZACH-ViT also gives the lowest unknown-attack BPCER at an APCER limit of 10\% ($81.29 \pm 1.95\%$). The high absolute errors show that the comparative advantage of the best compact model does not constitute deployment readiness under unknown PAIs.
\end{abstract}

\noindent\textbf{Keywords:} Iris presentation attack detection; biometric recognition security; compact vision transformers; ZACH-ViT; PAI shift; operational degradation; APCER; BPCER; D-EER.\par

\section{Introduction}
Iris recognition is widely used in biometric recognition because iris texture is distinctive and comparatively stable. However, iris recognition systems remain vulnerable to presentation attacks, including textured contact lenses and other artefacts designed to imitate bona-fide presentations. Presentation attack detection (PAD) is therefore a security-critical subsystem for person authentication rather than an auxiliary image-classification task~\cite{iso2382_37_2022,iso30107-1,yambay2017livdet}.

This paper studies a deployment-relevant challenge: a PAD model can appear adequate during development, yet degrade when the presentation attack instrument (PAI), sensor condition, or acquisition pipeline changes. A high Attack Presentation Classification Error Rate (APCER) under unknown attack presentations is a security-side error because attack presentations are classified as bona-fide presentations and may pass the PAD subsystem. A high Bona Fide Presentation Classification Error Rate (BPCER) affects availability and usability. Thus, the relevant question is not only which compact model obtains the best average result, but how the PAD error trade-off changes when the PAI, capture conditions, or operational scenario changes.

The LivDet-Iris 2017 Notre Dame subset supports this analysis through separate known-attack and unknown-attack test partitions~\cite{yambay2017livdet}. We interpret the known-to-unknown transition as PAI-driven domain shift. We additionally use light image corruptions as controlled proxies for sensor noise, focus degradation, optical contamination, compression, and low-quality capture pipelines. The evaluation therefore links compact model design, validation-driven threshold selection, and PAD security margins under shifted conditions.

We frame the study as a compact computer-vision benchmark rather than a single-architecture demonstration. Patch-ABMIL, Compact-TransMIL, and ZACH-ViT are evaluated under the same scratch-trained protocol, with every model containing at most approximately 0.26 million trainable parameters. Patch-ABMIL and Compact-TransMIL are project-specific compact variants inspired by attention-based deep multiple instance learning and TransMIL, respectively. ZACH-ViT is included as a zero-token compact vision transformer designed to reduce fixed spatial assumptions in data regimes where spatial organization is not uniformly informative~\cite{zachvit_lus,zachvit}.

\subsection{Contributions}
This paper makes four contributions:
\begin{enumerate}
    \item We present a compact iris PAD benchmark of scratch-trained computer-vision models containing at most approximately 0.26 million trainable parameters on LivDet-Iris 2017 Notre Dame.
    \item We quantify known-to-unknown security degradation, showing APCER increases of 17.11-30.47 percentage points and D-EER increases of 7.38-12.73 percentage points.
    \item We show that ZACH-ViT obtains the lowest unknown-attack and pooled-test APCER and D-EER at the validation-selected threshold, together with the lowest BPCER at an APCER limit of 10\%.
    \item We analyze architectural sensitivity and operational robustness through a ZACH-ViT ablation and controlled Gaussian-noise, blur, and Joint Photographic Experts Group (JPEG) compression stress tests.
\end{enumerate}

\section{Related Work}
Iris PAD spans textured contact lenses, printed or artificial eyes, and synthetic iris imagery, and generalization across PAIs remains open~\cite{czajka2018survey}. Doyle and Bowyer showed that robust textured contact lens detection can be achieved using Binarized Statistical Image Features (BSIF) descriptors without highly accurate iris segmentation~\cite{doyle2015bsif}, while LivDet-Iris established known-attack and unknown-attack protocols~\cite{yambay2017livdet}. Generative iris research has progressed from iDCGAN and RaSGAN to multi-domain CIT-GAN and diffusion-GAN synthesis, supporting both new PAIs and data augmentation~\cite{kohli2017idcgan,yadav2019rasgan,yadav2021citgan,yadav2025midstylegan}; LivDet-Iris 2023 further emphasizes synthetic PAIs~\cite{tinsley2023livdet}.

Deep iris PAD increasingly uses learned representations, including larger backbones, transfer learning, periocular information, and foundation models~\cite{tapia2025foundation}; privacy-preserving synthetic training is another emerging direction~\cite{mitcheff2024privacysafe}. Segmentation-aware methods have incorporated masks or joint iris localization~\cite{hoffman2018crosssensor,chen2018multitask}. Compact ocular recognition has also been evaluated on mobile hardware, while tiny-ML research highlights memory- and latency-aware architecture design~\cite{almadan2021ondevice,lin2020mcunet}. In contrast, we study compact scratch-trained models under PAI shift. ZACH-ViT first appeared in lung ultrasound~\cite{zachvit_lus}, was formalized as a regime-dependent inductive-bias study~\cite{zachvit}, and was subsequently evaluated under corruption and adversarial stress~\cite{zachvit_robust}.

\section{Experimental Protocol}

\subsection{Dataset and Evaluation Protocol}

We evaluate the Notre Dame subset of LivDet-Iris 2017, which contains an official
training partition and two official test partitions: known attack presentations and unknown
attack presentations~\cite{yambay2017livdet}. The data comprise near-infrared iris images
with bona-fide presentations and textured contact lens attacks derived from the Notre Dame
Contact Lenses Dataset 2015 lineage~\cite{doyle2015bsif}. The local manifest contains
1,200 training images, 1,800 known-attack test images, and 1,800 unknown-attack test images.

The official training set is split into 960 training and 240 validation images using a
stratified 80/20 split. All models are trained from scratch on the same split and evaluated
over seeds $\{3,5,7,11,13\}$. Repeated seeds quantify stochastic optimization variability
rather than class-subsampling variability. Each Tagged Image File Format (TIFF) image is
converted to grayscale, center-cropped, resized to $224 \times 224$, and replicated to three
channels. Center cropping avoids a separate segmentation model but retains non-iris context.
No external pretraining or data augmentation is used.

Terminology follows ISO/IEC 2382-37:2022~\cite{iso2382_37_2022}: we use
\emph{bona-fide presentation}, \emph{biometric presentation attack},
\emph{presentation attack detection} (PAD), and \emph{presentation attack instrument}
(PAI), avoiding legacy terms such as ``live'', ``fake'', or ``spoof''. Reporting follows
ISO/IEC 30107-style terminology~\cite{iso30107-1} and biometric performance-testing
practice~\cite{iso19795_1_2021}. We report Attack Presentation Classification Error Rate
(APCER), Bona Fide Presentation Classification Error Rate (BPCER), Detection Equal Error
Rate (D-EER), Detection Error Trade-off (DET) curves, and BPCER at a fixed APCER limit.

For the primary APCER and BPCER evaluation, the threshold is selected on validation data
using the APCER-BPCER balance criterion implemented in the evaluation scripts. It is then
transferred unchanged to the known-attack, unknown-attack, corrupted, and pooled test
partitions, avoiding oracle tuning of the primary test operating point. D-EER and DET
curves are computed by sweeping the threshold separately on each evaluation partition.
To characterize security-constrained operation, we additionally report BPCER at
APCER $\leq 10\%$, obtained from each test DET curve. This curve-derived operating
point characterizes the security-usability trade-off and does not replace or retune the
validation-selected threshold.

\subsection{Compact Baselines}   

The benchmark contains three compact scratch-trained models. Patch-ABMIL ($\sim$0.09 million parameters) is our patch-level compact variant inspired by attention-based deep multiple instance learning~\cite{ilse2018abmil}: non-overlapping image patches are treated as instances and aggregated with learned attention weights. Compact-TransMIL ($\sim$0.26 million) is our reduced transformer-MIL variant inspired by TransMIL~\cite{shao2021transmil}; unlike canonical TransMIL, it omits PPEG and Nystr\"om attention and uses a compact transformer stack followed by global average pooling. ZACH-ViT ($\sim$0.25 million) follows the previously introduced compact architecture~\cite{zachvit_lus,zachvit}. All models are trained for 23 epochs with batch size 16, the Adam optimizer, learning rate $10^{-4}$, and identical input size. The training schedule is fixed a priori across models to isolate architectural and threshold-transfer behavior rather than tune each model separately.

\section{Results}
\subsection{Main test performance}
Table~\ref{tab:main_results} summarizes performance at the validation-selected threshold and at the curve-derived APCER limit. On known attack presentations, ZACH-ViT obtains the lowest APCER and D-EER, while Compact-TransMIL has the lowest BPCER. Under unknown PAIs, ZACH-ViT gives the lowest APCER ($47.69 \pm 4.84\%$) and D-EER ($38.87 \pm 0.93\%$), while Compact-TransMIL has the lowest BPCER at the transferred threshold. ZACH-ViT also gives the lowest BPCER at APCER $\leq 10\%$ on the known-attack ($63.29 \pm 7.32\%$), unknown-attack ($81.29 \pm 1.95\%$), and pooled ($74.83 \pm 3.42\%$) partitions. Nevertheless, the unknown-attack result means that more than four-fifths of bona-fide presentations would be rejected when APCER is constrained to 10\%, indicating comparative advantage rather than operational acceptability.

\begin{figure}[htbp]
\centering
\begin{subfigure}{0.32\textwidth}
\includegraphics[width=\linewidth]{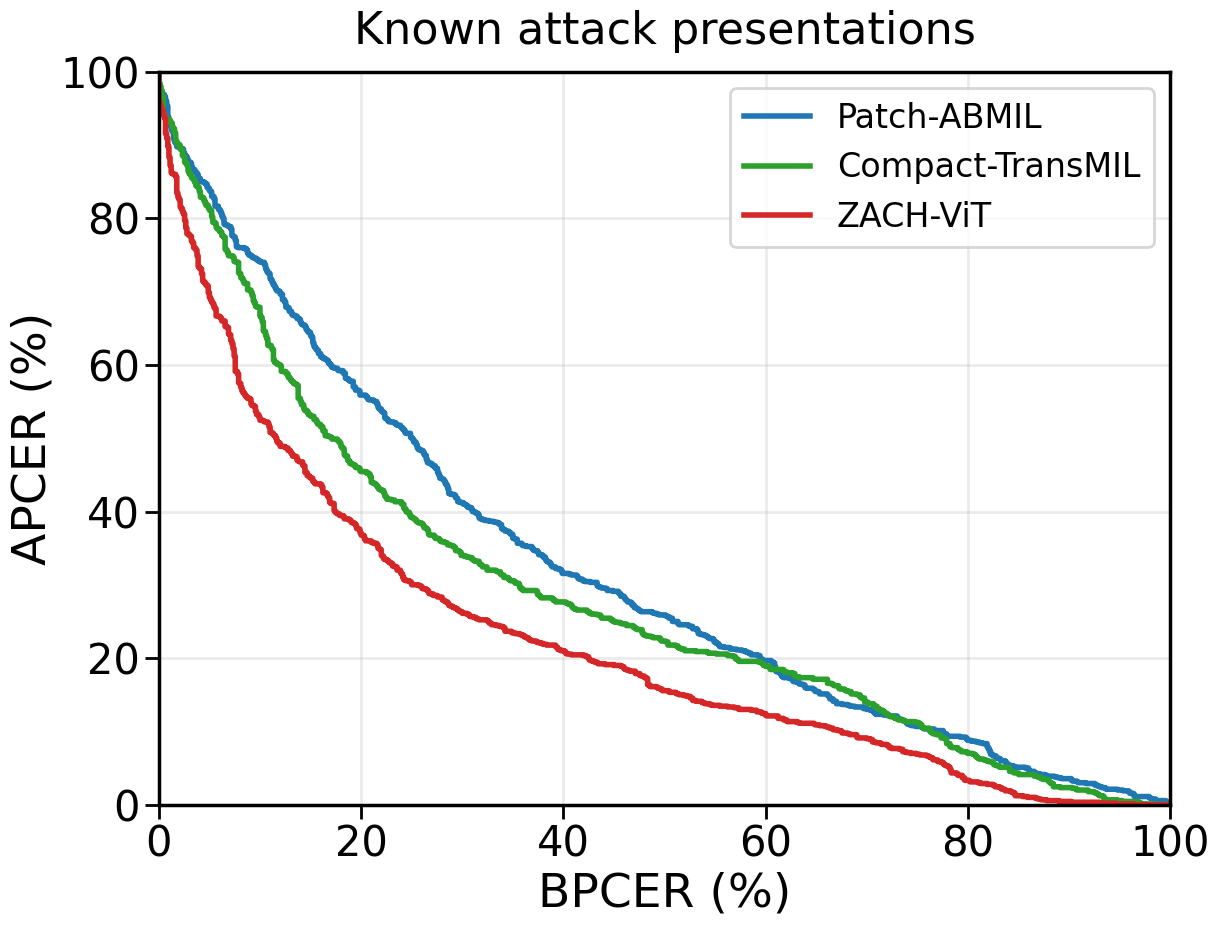}
\caption{Known attacks}
\end{subfigure}\hfill
\begin{subfigure}{0.32\textwidth}
\includegraphics[width=\linewidth]{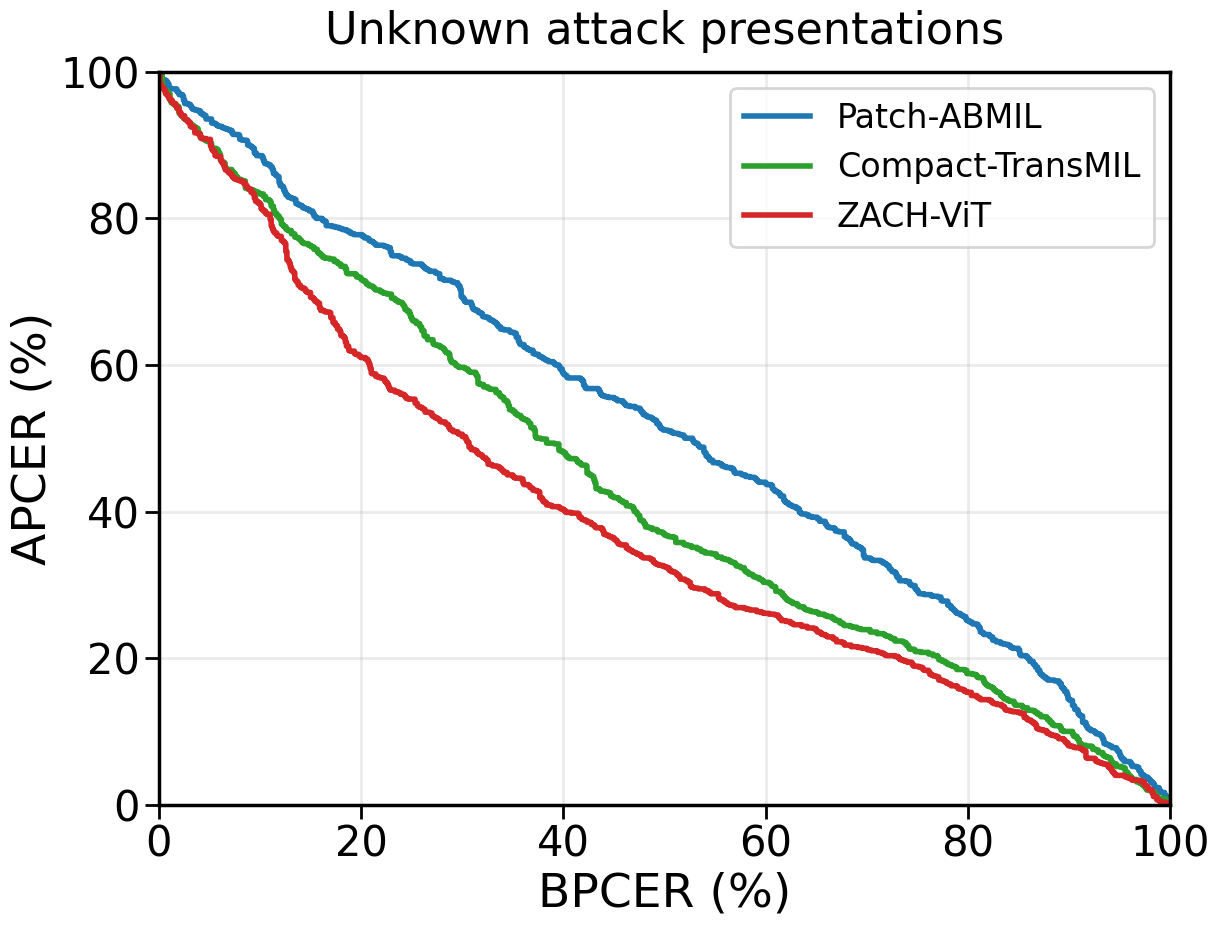}
\caption{Unknown attacks}
\end{subfigure}\hfill
\begin{subfigure}{0.32\textwidth}
\includegraphics[width=\linewidth]{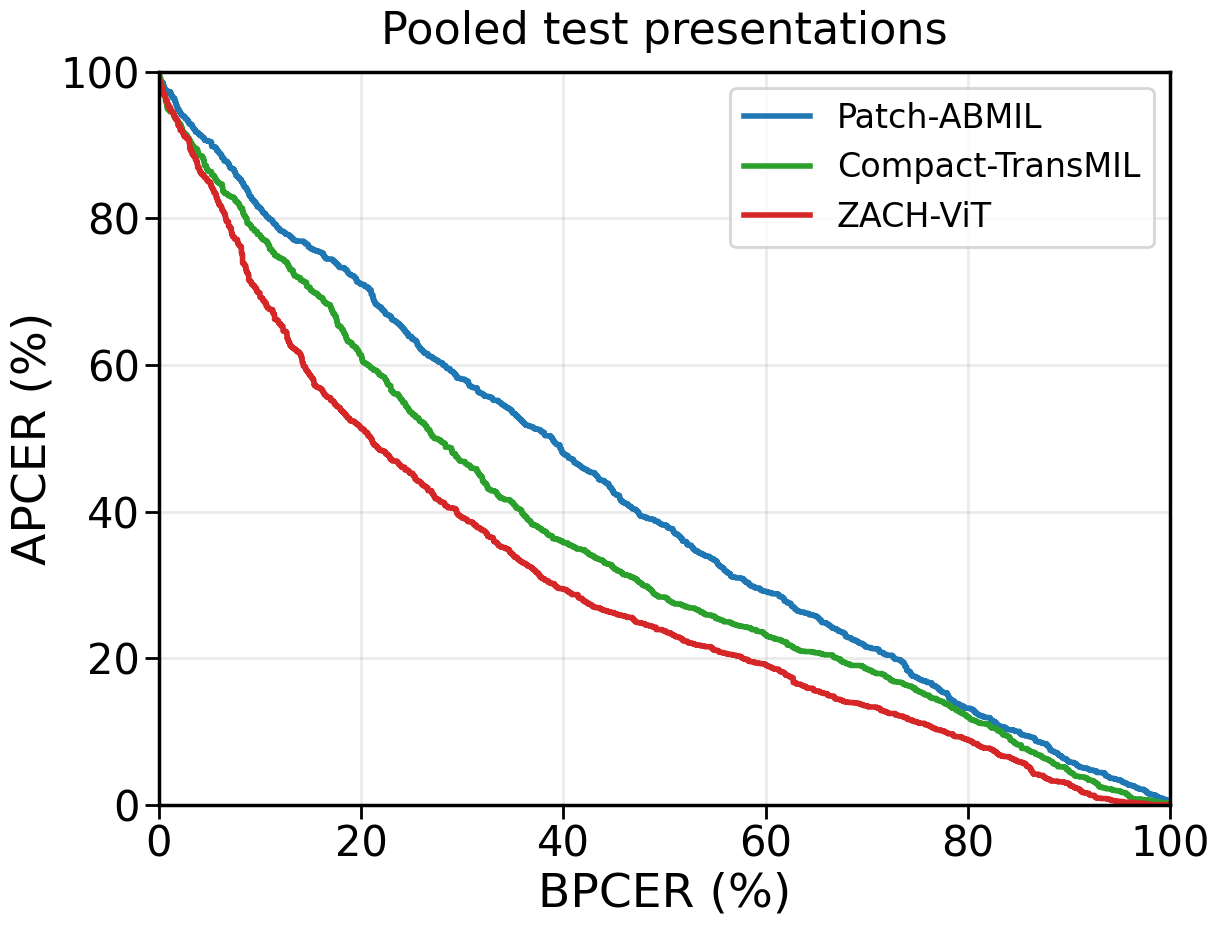}
\caption{Pooled test}
\end{subfigure}
\caption{Representative DET curves for seed 3. The horizontal axis is BPCER and the vertical axis is APCER.}
\label{fig:detcurves}
\end{figure}

Figure~\ref{fig:detcurves} shows representative DET curves for seed 3 and visualizes the BPCER-APCER trade-off across all test partitions.

\subsection{Known-to-unknown degradation}
Table~\ref{tab:degradation} quantifies degradation from known to unknown attack presentations. APCER increases by 17.11-30.47 percentage points and D-EER increases by 7.38-12.73 percentage points. ZACH-ViT has the smallest increase in both metrics and the lowest absolute unknown-attack APCER and D-EER. Its APCER rises from 30.58\% to 47.69\%, meaning that nearly half of unknown attack presentations are classified as bona-fide presentations at the validation-selected threshold. Compact-TransMIL has the lowest BPCER at that threshold.

\subsection{Ablation and corruption stress}
The ablation in Table~\ref{tab:ablation} tests positional embeddings, a classification (CLS) token, and patch size $8 \times 8$. The patch-size-8 variant gives the lowest known-attack APCER, BPCER, and D-EER, but generalizes poorly to unknown attack presentations, where APCER rises to $68.49 \pm 3.56\%$ and D-EER to $41.29 \pm 3.00\%$. On unknown attack presentations, the positional-embedding variant gives the lowest APCER, the patch-size-8 variant gives the lowest BPCER, and the default configuration gives the lowest D-EER. Strong performance on familiar PAIs therefore does not guarantee generalization to unseen PAIs.

Table~\ref{tab:corruptions} reports corruption-averaged results under Gaussian noise, Gaussian blur, and JPEG compression at two severities each. ZACH-ViT gives the lowest APCER, BPCER, and D-EER across the known-attack, unknown-attack, and pooled partitions. Nevertheless, its corrupted unknown-attack APCER remains $48.51 \pm 8.13\%$, representing a comparative robustness advantage rather than deployment certification.

\subsection{Efficiency}
Table~\ref{tab:efficiency} reports parameter count, training time, inference latency, throughput, and pooled-test PAD performance measured in the same local environment; these are comparative workstation measurements, not an on-device benchmark. Patch-ABMIL is the smallest and fastest model but has the weakest pooled-test performance. ZACH-ViT gives the lowest pooled-test BPCER at APCER $\leq 10\%$ and the lowest D-EER among models containing at most approximately 0.26 million trainable parameters.

\begin{table}[htbp]
\centering
\caption{Main PAD results on LivDet-Iris 2017 Notre Dame (mean $\pm$ standard deviation over five seeds). APCER and BPCER use the validation-selected threshold. D-EER and BPCER at APCER $\leq 10\%$ are derived by threshold sweeping. All values are percentages; lowest values per block and metric are bold.}
\label{tab:main_results}
\small
\begin{tabular}{lcccc}
\toprule
Model & APCER & BPCER & D-EER & BPCER@APCER$\leq10\%$ \\
\midrule
\multicolumn{5}{l}{\textit{Known attack presentations}} \\
Patch-ABMIL & $36.04 \pm 5.46$ & $36.09 \pm 3.83$ & $35.98 \pm 1.11$ & $72.02 \pm 4.95$ \\
Compact-TransMIL & $34.82 \pm 4.38$ & $\mathbf{30.91 \pm 2.44}$ & $32.67 \pm 2.82$ & $67.16 \pm 6.47$ \\
ZACH-ViT & $\mathbf{30.58 \pm 3.06}$ & $32.36 \pm 2.19$ & $\mathbf{31.49 \pm 2.05}$ & $\mathbf{63.29 \pm 7.32}$ \\
\addlinespace
\multicolumn{5}{l}{\textit{Unknown attack presentations}} \\
Patch-ABMIL & $66.51 \pm 3.74$ & $34.49 \pm 3.64$ & $48.71 \pm 2.33$ & $86.67 \pm 1.71$ \\
Compact-TransMIL & $53.62 \pm 3.62$ & $\mathbf{30.89 \pm 1.86}$ & $40.87 \pm 1.83$ & $81.96 \pm 3.32$ \\
ZACH-ViT & $\mathbf{47.69 \pm 4.84}$ & $31.84 \pm 2.59$ & $\mathbf{38.87 \pm 0.93}$ & $\mathbf{81.29 \pm 1.95}$ \\
\addlinespace
\multicolumn{5}{l}{\textit{Pooled test presentations}} \\
Patch-ABMIL & $51.28 \pm 4.07$ & $35.29 \pm 3.72$ & $43.02 \pm 1.26$ & $81.18 \pm 1.60$ \\
Compact-TransMIL & $44.22 \pm 2.83$ & $\mathbf{30.90 \pm 2.11}$ & $36.74 \pm 1.70$ & $76.90 \pm 3.98$ \\
ZACH-ViT & $\mathbf{39.13 \pm 2.78}$ & $32.10 \pm 2.07$ & $\mathbf{35.21 \pm 0.41}$ & $\mathbf{74.83 \pm 3.42}$ \\
\bottomrule
\end{tabular}
\end{table}

\begin{table}[htbp]
\centering
\caption{Known-to-unknown PAD degradation. Changes are unknown minus known in percentage points. Lowest absolute APCER and D-EER values and their smallest increases are bold. BPCER changes are descriptive because a lower BPCER under shift may coincide with a higher APCER.}
\label{tab:degradation}
\small
\begin{tabular}{llrrr}
\toprule
Metric & Model & Known & Unknown & Change \\
\midrule
APCER & Patch-ABMIL & 36.04 & 66.51 & 30.47 \\
 & Compact-TransMIL & 34.82 & 53.62 & 18.80 \\
 & ZACH-ViT & $\mathbf{30.58}$ & $\mathbf{47.69}$ & $\mathbf{17.11}$ \\
\addlinespace
BPCER & Patch-ABMIL & 36.09 & 34.49 & -1.60 \\
 & Compact-TransMIL & 30.91 & 30.89 & -0.02 \\
 & ZACH-ViT & 32.36 & 31.84 & -0.51 \\
\addlinespace
D-EER & Patch-ABMIL & 35.98 & 48.71 & 12.73 \\
 & Compact-TransMIL & 32.67 & 40.87 & 8.20 \\
 & ZACH-ViT & $\mathbf{31.49}$ & $\mathbf{38.87}$ & $\mathbf{7.38}$ \\
\bottomrule
\end{tabular}
\end{table}

\begin{table}[htbp]
\centering
\caption{PAD ablation of ZACH-ViT (mean $\pm$ standard deviation over five seeds, in percent). PosEmb denotes positional embeddings and PS8 denotes patch size $8 \times 8$. Lowest values per block and metric are bold.}
\label{tab:ablation}
\small
\begin{tabular}{lccc}
\toprule
Variant & APCER & BPCER & D-EER \\
\midrule
\multicolumn{4}{l}{\textit{Known attack presentations}} \\
Default & $30.58 \pm 3.06$ & $32.36 \pm 2.19$ & $31.49 \pm 2.05$ \\
+ PosEmb & $29.11 \pm 2.46$ & $36.64 \pm 1.47$ & $33.11 \pm 0.86$ \\
+ CLS & $35.13 \pm 4.90$ & $32.87 \pm 4.22$ & $33.69 \pm 1.15$ \\
PS8 & $\mathbf{25.07 \pm 3.38}$ & $\mathbf{24.40 \pm 2.80}$ & $\mathbf{24.87 \pm 3.00}$ \\
\addlinespace
\multicolumn{4}{l}{\textit{Unknown attack presentations}} \\
Default & $47.69 \pm 4.84$ & $31.84 \pm 2.59$ & $\mathbf{38.87 \pm 0.93}$ \\
+ PosEmb & $\mathbf{43.58 \pm 3.02}$ & $35.33 \pm 0.94$ & $39.16 \pm 1.06$ \\
+ CLS & $50.71 \pm 6.26$ & $33.69 \pm 4.12$ & $41.49 \pm 1.31$ \\
PS8 & $68.49 \pm 3.56$ & $\mathbf{23.22 \pm 3.16}$ & $41.29 \pm 3.00$ \\
\addlinespace
\multicolumn{4}{l}{\textit{Pooled test presentations}} \\
Default & $39.13 \pm 2.78$ & $32.10 \pm 2.07$ & $35.21 \pm 0.41$ \\
+ PosEmb & $\mathbf{36.34 \pm 1.30}$ & $35.99 \pm 1.02$ & $36.22 \pm 0.62$ \\
+ CLS & $42.92 \pm 5.40$ & $33.28 \pm 4.08$ & $37.43 \pm 0.39$ \\
PS8 & $46.78 \pm 3.02$ & $\mathbf{23.81 \pm 2.97}$ & $\mathbf{33.72 \pm 2.79}$ \\
\bottomrule
\end{tabular}
\end{table}

\begin{table}[htbp]
\centering
\caption{Corruption-averaged robustness under Gaussian noise, Gaussian blur, and JPEG compression (mean $\pm$ standard deviation over corruption, severity, and seed evaluations, in percent). Lowest values per block and metric are bold.}
\label{tab:corruptions}
\small
\begin{tabular}{lccc}
\toprule
Model & APCER & BPCER & D-EER \\
\midrule
\multicolumn{4}{l}{\textit{Known attack presentations}} \\
Patch-ABMIL & $37.19 \pm 5.53$ & $35.66 \pm 3.82$ & $36.29 \pm 1.03$ \\
Compact-TransMIL & $33.93 \pm 5.75$ & $34.37 \pm 5.68$ & $34.20 \pm 2.94$ \\
ZACH-ViT & $\mathbf{32.10 \pm 7.71}$ & $\mathbf{32.96 \pm 4.80}$ & $\mathbf{32.48 \pm 2.00}$ \\
\addlinespace
\multicolumn{4}{l}{\textit{Unknown attack presentations}} \\
Patch-ABMIL & $67.50 \pm 3.87$ & $33.82 \pm 3.27$ & $48.54 \pm 2.25$ \\
Compact-TransMIL & $52.08 \pm 6.47$ & $33.88 \pm 5.58$ & $42.09 \pm 1.99$ \\
ZACH-ViT & $\mathbf{48.51 \pm 8.13}$ & $\mathbf{32.76 \pm 4.49}$ & $\mathbf{39.84 \pm 1.71}$ \\
\addlinespace
\multicolumn{4}{l}{\textit{Pooled test presentations}} \\
Patch-ABMIL & $52.38 \pm 4.30$ & $34.72 \pm 3.48$ & $42.98 \pm 1.12$ \\
Compact-TransMIL & $43.01 \pm 5.52$ & $34.19 \pm 5.68$ & $38.25 \pm 2.34$ \\
ZACH-ViT & $\mathbf{40.20 \pm 7.32}$ & $\mathbf{32.86 \pm 4.63}$ & $\mathbf{36.25 \pm 1.46}$ \\
\bottomrule
\end{tabular}
\end{table}

\begin{table}[htbp]
\centering
\caption{Efficiency and deployment-oriented summary. FPS denotes frames per second. BPCER at APCER $\leq 10\%$ is derived from the pooled-test DET curve. Lowest BPCER and D-EER are bold.}
\label{tab:efficiency}
\small
\setlength{\tabcolsep}{2.2pt}
\begin{tabular}{lcccccc}
\toprule
Model & Par. (M) & Train (s) & Lat. (ms) & FPS & BPCER@10\% APCER & D-EER \\
\midrule
ZACH-ViT & 0.25 & $82.11 \pm 1.36$ & $1.123 \pm 0.028$ & 890.33 & $\mathbf{74.83 \pm 3.42}$ & $\mathbf{35.21 \pm 0.41}$ \\
Compact-TransMIL & 0.26 & $81.80 \pm 3.08$ & $1.085 \pm 0.115$ & 921.47 & $76.90 \pm 3.98$ & $36.74 \pm 1.70$ \\
Patch-ABMIL & 0.09 & $72.15 \pm 0.79$ & $0.276 \pm 0.012$ & 3623.89 & $81.18 \pm 1.60$ & $43.02 \pm 1.26$ \\
\bottomrule
\end{tabular}
\end{table}

\section{Discussion}

Compact computer-vision models show substantial degradation when the PAI changes.
From known to unknown attack presentations, APCER increases by 17.11--30.47
percentage points and D-EER by 7.38--12.73 percentage points. ZACH-ViT gives the
lowest unknown-attack APCER and D-EER, whereas Compact-TransMIL gives the lowest BPCER
at the validation-selected threshold.

The threshold protocol is central to this interpretation. APCER and BPCER at the
transferred threshold describe a fixed decision rule, whereas D-EER, DET curves, and
BPCER at a fixed APCER describe the score-distribution trade-off. At APCER
$\leq 10\%$, unknown-attack BPCER remains 81.29\%--86.67\%; at APCER limits of
1\% and 0.10\%, it rises above 96.5\% and 99.7\%, respectively. Thus, none of the
evaluated models provides an acceptable security-usability operating point under
unknown PAIs.

The ablation results further show that stronger known-attack performance does not
guarantee generalization. Reducing patch size improves results for familiar PAIs but
weakens unknown-attack performance, suggesting over-reliance on local texture.
Similarly, noise, blur, and compression should be evaluated using the threshold selected
on clean validation data rather than by retuning on corrupted test samples.

Center cropping avoids iris-segmentation latency and failures but retains sclera, eyelids,
periocular texture, and acquisition background. Because ZACH-ViT uses global average
pooling, this additional context may contain useful PAI evidence or dataset-specific
shortcuts. Comparing center-cropped, iris-masked, and segmentation-aware inputs is
therefore an important next step~\cite{hoffman2018crosssensor,chen2018multitask}.
Compactness must also be evaluated together with security behavior: Patch-ABMIL is the
smallest and fastest model but has the highest unknown-attack APCER. Actual edge
deployment should additionally measure memory, energy, and latency on target
hardware~\cite{almadan2021ondevice,lin2020mcunet}.

\section{Conclusion}

We evaluated compact scratch-trained models for iris PAD under PAI-driven domain
shift and environmental degradation. ZACH-ViT gives the lowest unknown-attack and
pooled-test APCER and D-EER, as well as the lowest unknown-attack BPCER at APCER
$\leq 10\%$ ($81.29 \pm 1.95\%$). It is therefore the strongest comparative compact
model in this benchmark, but the absolute error rates do not support deployment under
unknown PAIs.

The study is limited to one benchmark, center-crop preprocessing, workstation-level
efficiency measurements, and simple corruption tests. Future work should compare
segmentation strategies, evaluate recent synthetic PAIs, cross-sensor and temporal
conditions, adaptive threats, and on-device performance, while reporting known and
unknown PAIs separately under both transferred and security-constrained operating
points.

\printbibliography

@manual{iso2382_37_2022,
  title        = {{Information technology, Vocabulary, Part 37: Biometrics}},
  organization = {International Organization for Standardization and International Electrotechnical Commission},
  number       = {ISO/IEC 2382-37:2022},
  address      = {Geneva, Switzerland},
  year         = {2022}
}

@manual{iso30107-1,
  title        = {{Information technology, Biometric presentation attack detection, Part 1: Framework}},
  organization = {International Organization for Standardization},
  number       = {ISO/IEC 30107-1:2016},
  address      = {Geneva, Switzerland},
  year         = {2016}
}

@manual{iso19795_1_2021,
  title        = {{Information technology, Biometric performance testing and reporting, Part 1: Principles and framework}},
  organization = {International Organization for Standardization and International Electrotechnical Commission},
  number       = {ISO/IEC 19795-1:2021},
  address      = {Geneva, Switzerland},
  year         = {2021}
}

@inproceedings{yambay2017livdet,
  author    = {Yambay, David and Becker, Brian and Kohli, Naman and Czajka, Adam and Bowyer, Kevin W. and Schuckers, Stephanie and Singh, Richa and Vatsa, Mayank and Noore, Afzel and Tan, Tieniu},
  title     = {{LivDet Iris 2017 - Iris Liveness Detection Competition 2017}},
  booktitle = {Proceedings of the IEEE International Joint Conference on Biometrics (IJCB)},
  pages     = {733--741},
  year      = {2017},
  doi       = {10.1109/BTAS.2017.8272763}
}

@misc{zachvit_lus,
  author       = {Angelakis, Athanasios and others},
  title        = {{ZACH-ViT: A Zero-Token Vision Transformer with ShuffleStrides Data Augmentation for Robust Lung Ultrasound Classification}},
  howpublished = {arXiv preprint arXiv:2510.17650},
  year         = {2025}
}

@misc{zachvit,
  author       = {Angelakis, Athanasios},
  title        = {{ZACH-ViT: Regime-Dependent Inductive Bias in Compact Vision Transformers for Medical Imaging}},
  howpublished = {arXiv preprint arXiv:2602.17929},
  year         = {2026},
  doi          = {10.48550/arXiv.2602.17929}
}

@inproceedings{zachvit_robust,
  author    = {Angelakis, Athanasios and Gomez-Barrero, Marta},
  title     = {{Extending ZACH-ViT to Robust Medical Imaging: Corruption and Adversarial Stress Testing in Low-Data Regimes}},
  booktitle = {Proceedings of the IEEE/CVF Conference on Computer Vision and Pattern Recognition Workshops (CVPRW)},
  pages     = {6114--6122},
  year      = {2026}
}

@article{doyle2015bsif,
  author  = {Doyle, James S. and Bowyer, Kevin W.},
  title   = {{Robust Detection of Textured Contact Lenses in Iris Recognition Using BSIF}},
  journal = {IEEE Access},
  volume  = {3},
  pages   = {1672--1683},
  year    = {2015},
  doi     = {10.1109/ACCESS.2015.2477470}
}

@misc{tinsley2023livdet,
  author       = {Tinsley, Patrick and Purnapatra, Sourav and Mitcheff, Mary and Boyd, Andrew and Crum, Claire and Bowyer, Kevin W. and Flynn, Patrick and Schuckers, Stephanie and Czajka, Adam and Fang, Meiling and Damer, Naser and Liu, Xin and Wang, Cunjian and Sun, Xintao and Chang, Zhenan and Li, Xiaoming and Zhao, Guodong and Tapia, Juan E. and Busch, Christoph and Aravena, Claudio and Schulz, Daniel},
  title        = {{Iris Liveness Detection Competition (LivDet-Iris)}},
  howpublished = {arXiv preprint arXiv:2310.04541},
  year         = {2023}
}

@misc{tapia2025foundation,
  author       = {Tapia, Juan E. and Gonz\'alez-Soler, L. J. and Busch, Christoph},
  title        = {{Towards Iris Presentation Attack Detection with Foundation Models}},
  howpublished = {arXiv preprint arXiv:2501.06312},
  year         = {2025}
}

@inproceedings{mitcheff2024privacysafe,
  author    = {Mitcheff, Mary and Boyd, Andrew and Tinsley, Patrick and Czajka, Adam and Bowyer, Kevin W. and Flynn, Patrick J.},
  title     = {{Privacy-Safe Iris Presentation Attack Detection}},
  booktitle = {Proceedings of the IEEE International Joint Conference on Biometrics (IJCB)},
  year      = {2024}
}

@inproceedings{ilse2018abmil,
  author    = {Ilse, Maximilian and Tomczak, Jakub M. and Welling, Max},
  title     = {{Attention-based Deep Multiple Instance Learning}},
  booktitle = {Proceedings of the 35th International Conference on Machine Learning (ICML)},
  volume    = {80},
  pages     = {2127--2136},
  year      = {2018},
  doi       = {10.48550/arXiv.1802.04712}
}

@inproceedings{shao2021transmil,
  author    = {Shao, Zhuchen and Bian, Hao and Chen, Yang and Wang, Yifeng and Zhang, Jian and Ji, Xiangyang and Zhang, Yongbing},
  title     = {{TransMIL: Transformer Based Correlated Multiple Instance Learning for Whole Slide Image Classification}},
  booktitle = {Advances in Neural Information Processing Systems 34 (NeurIPS)},
  pages     = {2136--2147},
  year      = {2021},
  doi       = {10.48550/arXiv.2106.00908}
}

@article{czajka2018survey,
  author  = {Czajka, Adam and Bowyer, Kevin W.},
  title   = {{Presentation Attack Detection for Iris Recognition: An Assessment of the State-of-the-Art}},
  journal = {ACM Computing Surveys},
  volume  = {51},
  number  = {4},
  pages   = {86:1--86:35},
  year    = {2018},
  doi     = {10.1145/3232849}
}

@inproceedings{kohli2017idcgan,
  author    = {Kohli, Naman and Yadav, Daksha and Vatsa, Mayank and Singh, Richa and Noore, Afzel},
  title     = {{Synthetic Iris Presentation Attack Using iDCGAN}},
  booktitle = {Proceedings of the IEEE International Joint Conference on Biometrics (IJCB)},
  pages     = {674--680},
  year      = {2017},
  doi       = {10.1109/BTAS.2017.8272756}
}

@inproceedings{yadav2019rasgan,
  author    = {Yadav, Shivangi and Chen, Cunjian and Ross, Arun},
  title     = {{Synthesizing Iris Images Using RaSGAN With Application in Presentation Attack Detection}},
  booktitle = {Proceedings of the IEEE/CVF Conference on Computer Vision and Pattern Recognition Workshops (CVPRW)},
  pages     = {2422--2430},
  year      = {2019},
  doi       = {10.1109/CVPRW.2019.00297}
}

@inproceedings{yadav2021citgan,
  author    = {Yadav, Shivangi and Ross, Arun},
  title     = {{CIT-GAN: Cyclic Image Translation Generative Adversarial Network With Application in Iris Presentation Attack Detection}},
  booktitle = {Proceedings of the IEEE/CVF Winter Conference on Applications of Computer Vision (WACV)},
  pages     = {2412--2421},
  year      = {2021}
}

@inproceedings{yadav2025midstylegan,
  author    = {Yadav, Shivangi and Ross, Arun},
  title     = {{A Multi-domain Image Translative Diffusion StyleGAN for Iris Presentation Attack Detection}},
  booktitle = {Proceedings of the IEEE/CVF International Conference on Computer Vision Workshops (ICCVW)},
  pages     = {3747--3756},
  year      = {2025}
}

@inproceedings{hoffman2018crosssensor,
  author    = {Hoffman, Steven and Sharma, Renu and Ross, Arun},
  title     = {{Convolutional Neural Networks for Iris Presentation Attack Detection: Toward Cross-Dataset and Cross-Sensor Generalization}},
  booktitle = {Proceedings of the IEEE/CVF Conference on Computer Vision and Pattern Recognition Workshops (CVPRW)},
  pages     = {1620--1628},
  year      = {2018},
  doi       = {10.1109/CVPRW.2018.00213}
}

@inproceedings{chen2018multitask,
  author    = {Chen, Cunjian and Ross, Arun},
  title     = {{A Multi-task Convolutional Neural Network for Joint Iris Detection and Presentation Attack Detection}},
  booktitle = {Proceedings of the IEEE Winter Conference on Applications of Computer Vision Workshops (WACVW)},
  pages     = {44--51},
  year      = {2018},
  doi       = {10.1109/WACVW.2018.00011}
}

@inproceedings{almadan2021ondevice,
  author    = {Almadan, Ali and Rattani, Ajita},
  title     = {{Compact CNN Models for On-device Ocular-based User Recognition in Mobile Devices}},
  booktitle = {Proceedings of the IEEE Symposium Series on Computational Intelligence (SSCI)},
  pages     = {1--7},
  year      = {2021},
  doi       = {10.1109/SSCI50451.2021.9660033}
}

@inproceedings{lin2020mcunet,
  author    = {Lin, Ji and Chen, Wei-Ming and Lin, Yujun and Cohn, John and Gan, Chuang and Han, Song},
  title     = {{MCUNet: Tiny Deep Learning on IoT Devices}},
  booktitle = {Advances in Neural Information Processing Systems 33 (NeurIPS)},
  year      = {2020}
}

\end{document}